\documentclass{article}

\usepackage{header}

\usepackage[accepted,nohyperref]{icml2020}

\begin{document}

\twocolumn[
\icmltitle{REALM: Retrieval-Augmented Language Model Pre-Training}

\icmlsetsymbol{equal}{*}

\begin{icmlauthorlist}
\icmlauthor{Kelvin Guu}{equal,google}
\icmlauthor{Kenton Lee}{equal,google}
\icmlauthor{Zora Tung}{google}
\icmlauthor{Panupong Pasupat}{google}
\icmlauthor{Ming-Wei Chang}{google}
\end{icmlauthorlist}

\icmlaffiliation{google}{Google Research}
\icmlcorrespondingauthor{Kelvin Guu}{kguu@google.com}
\icmlcorrespondingauthor{Kenton Lee}{kentonl@google.com}
\icmlcorrespondingauthor{Zora Tung}{gatoatigrado@google.com}
\icmlcorrespondingauthor{Panupong Pasupat}{ppasupat@google.com}
\icmlcorrespondingauthor{Ming-Wei Chang}{mingweichang@google.com}

\icmlkeywords{Machine Learning, ICML, Neural Networks, Information Retrieval, Question Answering, Natural Language Processing}

\vskip 0.3in
]

\printAffiliationsAndNotice{\icmlEqualContribution} 

\begin{abstract}
Language model pre-training has been shown to capture a surprising amount of world knowledge, crucial for NLP tasks such as question answering. However, this knowledge is stored implicitly in the parameters of a neural network, requiring ever-larger networks to cover more facts.
To capture knowledge in a more modular and interpretable way, we augment language model pre-training with a latent {\em knowledge retriever}, which allows the model to retrieve and attend over documents from a large corpus such as Wikipedia, used during pre-training, fine-tuning and inference. For the first time, we show how to pre-train such a knowledge retriever in an unsupervised manner, using masked language modeling as the learning signal and backpropagating through a retrieval step that considers millions of documents.
We demonstrate the effectiveness of \thefullname pre-training (\thename) by fine-tuning on the challenging task of Open-domain Question Answering (\openqa). We compare against state-of-the-art models for both explicit and implicit knowledge storage on three popular \openqa benchmarks, and find that we outperform all previous methods by a significant margin (4-16\% absolute accuracy), while also providing qualitative benefits such as interpretability and modularity. \end{abstract}
\section{Introduction}

\begin{figure}[t]
\centering
\includegraphics[width=\columnwidth]{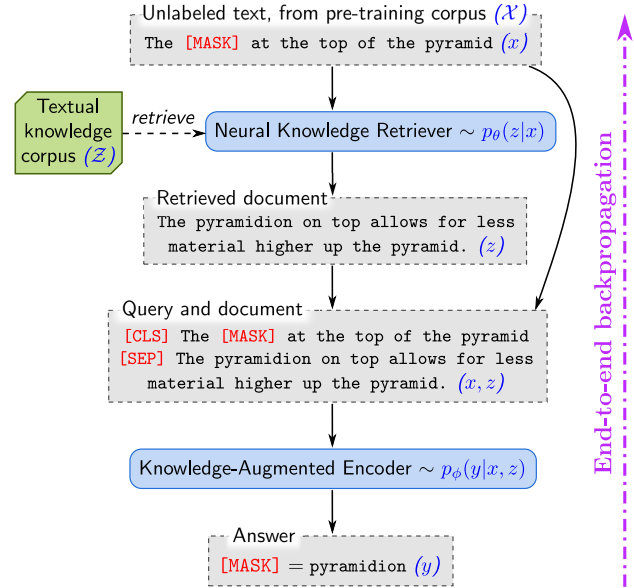}
\caption{\label{fig:intro} \ \thename augments language model pre-training with a {\bf neural knowledge retriever}
that retrieves knowledge from a {\bf textual knowledge corpus}, $\cZ$ (e.g., all of Wikipedia). Signal from the language modeling objective backpropagates all the way through the retriever, which must consider millions of documents in $\cZ$---a significant computational challenge that we address.}
\end{figure}

Recent advances in language model pre-training have shown that models such as BERT~\cite{bert}, RoBERTa~\cite{roberta} and T5~\cite{t5} store a surprising amount of world knowledge,
acquired from the massive text corpora they are trained on~\cite{lm_as_kb}.
For example, BERT is able to correctly predict the missing word in the following sentence:
\nl{The \blank\xspace is the currency of the United Kingdom} (answer: \nl{pound}).

In these language models, the learned world knowledge is stored {\em implicitly} in the parameters of the underlying neural network. This makes it difficult to determine what knowledge is stored in the network and where.
Furthermore, storage space is limited by the size of the network---to capture more world knowledge, one must train ever-larger networks, which can be prohibitively slow or expensive. 

To capture knowledge in a more interpretable and modular way, we propose a novel framework, \thefullname (\thename) pre-training, which augments language model pre-training algorithms with a learned {\em textual knowledge retriever}.
In contrast to models that store knowledge in their parameters, this approach \emph{explicitly} exposes the role of world knowledge by asking the model to decide what knowledge to retrieve and use during inference.
Before making each prediction, the language model uses the retriever to retrieve documents\footnotemark~from a large corpus such as Wikipedia,
and then attends over those documents to help inform its prediction.
Learning this model end-to-end requires backpropagating through a retrieval step that considers an entire corpus of textual knowledge, as shown in Figure~\ref{fig:intro}.

The key intuition of \thename is to
train the retriever using a {\em performance-based} signal from unsupervised text:
a retrieval that {\em improves} the language model's perplexity is helpful and should be rewarded, while an uninformative retrieval should be penalized. 
For example, in Figure~\ref{fig:intro}, if the model needs to fill the blank in \nl{the~\blank~at the top of the pyramid}, the retriever should be rewarded for selecting a document containing \nl{The pyramidion on top allows for less material higher up the pyramid}. We achieve this behavior by modeling our {\em retrieve-then-predict} approach as a latent variable language model and optimizing the marginal likelihood.

Incorporating a large-scale neural retrieval module
during pre-training constitutes a significant computational challenge, since the retriever must consider millions of candidate documents for each pre-training step, and we must backpropagate through its decisions. To address this, we structure the retriever such that the computation performed for each document can be cached and asynchronously updated, and selection of the best documents can be formulated as Maximum Inner Product Search (MIPS).

Numerous prior works have
demonstrated the benefit of adding a discrete retrieval step to neural networks~\cite{key_value_memorynetwork, drqa}, but did not apply the framework
to language model pre-training and employed non-learned retrievers to handle large-scale document collections.
In the language modeling literature, the $k$-Nearest Neighbor Language Model~\cite{knnlm} ($k$NN-LM) retrieves similar LM examples to improve memorization. However, $k$NN-LM was not fine-tuned for downstream tasks, perhaps because it is unclear how to adapt the retrieval mechanism: a $k$NN can only use examples labeled for the target task---during fine-tuning, this precludes LM examples, which contain the desired world knowledge. In contrast, \thename's retriever is designed to transfer to other tasks, and the retrieval is just text, not a labeled example.

We evaluate our approach by fine-tuning the models pre-trained with \thename on the task of Open-domain Question Answering (\openqa), one of the most knowledge-intensive tasks in natural language processing.
We evaluate on three popular \openqa benchmarks (\nq, \wq, and \trec) and compare to state-of-the-art \openqa models, including both extremely large models that store knowledge implicitly (such as T5) as well as previous approaches that also use a knowledge retriever to access external knowledge, but implement retrieval in a more heuristic fashion~\cite{orqa,openqa_hardem,rrp_salesforce}. \thename achieves new state-of-the-art results on all three benchmarks, significantly outperforming all previous systems by 4-16\% absolute accuracy. We also demonstrate qualitative benefits of \thename, including interpretability and modularity.

\footnotetext{We use the term ``document'' loosely to refer to a passage from the knowledge corpus, not necessarily a whole article.} \section{Background}
\label{sec:background}
\paragraph{Language model pre-training}

The goal of language model pre-training is to learn useful representations of language,
usually from unlabeled text corpora.
The resulting pre-trained model can then be further trained (\emph{fine-tuned}) for a downstream task of primary interest (in our case, \openqa),
often leading to better generalization than training from scratch \cite{dai_finetune,gpt2}.

We focus on the {\em masked language model}\footnote{Strictly speaking, MLM is not a standard language model, since it does not define a distribution over the entire sequence of tokens. In the paper we sometimes abuse the term ``language model'' slightly to make the phrase shorter.}
(MLM) variant of pre-training popularized by BERT~\cite{bert}.
In its basic form, an MLM is trained to predict the missing tokens in an input text passage.
Given an unlabeled pre-training corpus $\mathcal{X}$ (e.g., Wikipedia text),
a training example $(x, y)$ can be generated by randomly masking tokens in a sampled piece of text
(e.g., $x =$ \nl{The \mask is the currency \mask the UK}; $y =$ (\nl{pound}, \nl{of})).
The model uses its representation of the masked input $x$ to predict
the token that should go in each mask.
A good MLM must learn to encode syntactic and semantic information (e.g., to predict \nl{of})
as well as some world knowledge (e.g., to predict \nl{pound}).

\paragraph{Open-domain question answering (\openqa)}
To measure a model's ability to incorporate world knowledge, we need a downstream task where world knowledge is critical.
Perhaps one of the most knowledge-intensive tasks in natural language processing is open-domain question answering (\openqa):
given a question $x$ such as ``\texttt{\small What is the currency of the UK?}'', a model must output the correct answer string $y$, ``\texttt{{\small pound}}''.
The ``open'' part of \openqa refers to the fact that the model does {\em not} receive a pre-identified document that is known to contain the answer, unlike traditional reading comprehension (RC) tasks such as SQuAD \cite{squad, squad2}. While RC models comprehend a single document, \openqa models must retain knowledge from millions of documents, since a question could be about any of them.

We focus on \openqa systems that utilize a \emph{textual knowledge corpus} $\cZ$ as the knowledge source.
Many of these systems employ a \emph{retrieval-based} approach:
given a question $x$, retrieve potentially relevant documents $z$ from the corpus $\cZ$,
and then extract an answer $y$ from the documents~\cite{askmsr,drqa,orqa}.
Our approach, \thename, is inspired by this paradigm and extends it to language model pre-training.
Alternatively, some recent work has proposed \emph{generation-based} systems that
apply a sequence-to-sequence model on $x$ to directly generate $y$ token-by-token
\cite{bart_not_bert,t5}.
We will compare against state-of-the-art systems from both paradigms in our experiments. \section{Approach} \label{sec:approach}
\begin{figure*}[t!]
\centering
\includegraphics[width=.9\textwidth]{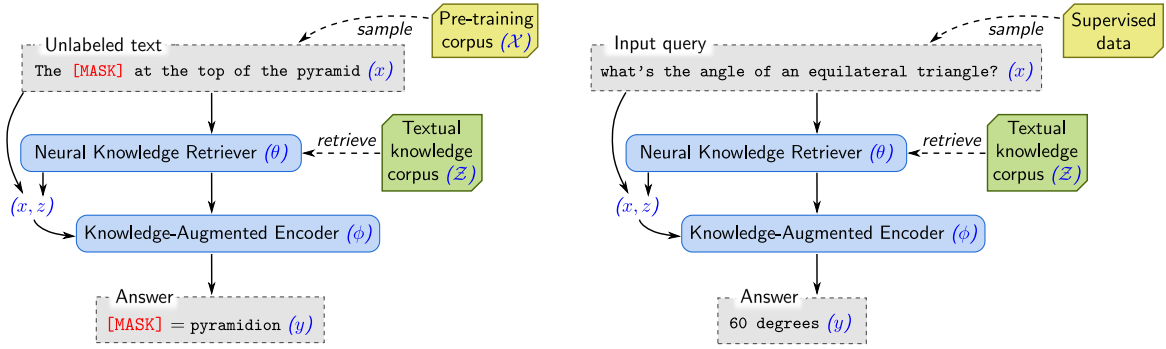}
\caption{The overall framework of \thename. {\bf Left:} {\em Unsupervised pre-training.} The knowledge retriever and knowledge-augmented encoder are jointly pre-trained on the unsupervised language modeling task. 
{\bf Right:} {\em Supervised fine-tuning.} After the parameters of the retriever ($\theta$) and encoder ($\phi$) have been pre-trained, they are then fine-tuned on a task of primary interest, using supervised examples.}
\label{fig:approach}
\end{figure*}

We start by formalizing \thename's pre-training and fine-tuning tasks as a {\em retrieve-then-predict} generative process in Section~\ref{sec:generative_process}. Then in Section~\ref{sec:model_architecture}, we describe the model architectures for each component of that process. In Section~\ref{sec:training}, we show how to implement \thename pre-training and fine-tuning by maximizing the likelihood of \thename's generative process. En route, we address important computational challenges, explain why training works, and also discuss strategies for injecting useful inductive biases. The overall framework is illustrated in Figure~\ref{fig:approach}.

\subsection{\thename's generative process}
\label{sec:generative_process}
For both pre-training and fine-tuning, \thename takes some input $x$ and learns a distribution $p(y \mid x)$ over possible outputs $y$. For pre-training, the task is masked language modeling: $x$ is a sentence from a pre-training corpus $\cX$ with some tokens masked out, and the model must predict the value of those missing tokens, $y$. For fine-tuning, the task is \openqa: $x$ is a question, and $y$ is the answer.

\thename decomposes $p(y\mid x)$ into two steps: {\em retrieve}, then {\em predict}. Given an input $x$, we first retrieve possibly helpful documents $z$ from a knowledge corpus $\cZ$. We model this as a sample from the distribution $p(z\mid x)$. Then, we condition on both the retrieved $z$ and the original input $x$ to generate the output $y$---modeled as $p(y\mid z,x)$. To obtain the overall likelihood of generating $y$, we treat $z$ as a latent variable and marginalize over all possible documents $z$, yielding \begin{equation}
p(y \mid x) = \sum_{z \in \cZ} p(y \mid z, x)\, p(z \mid x). \label{eqn:marginal}
\end{equation}

\subsection{Model architecture}
\label{sec:model_architecture}
We now describe the two key components: the \textbf{neural knowledge retriever}, which models $p(z\mid x)$, and the \textbf{knowledge-augmented encoder}, which models $p(y\mid z,x)$.

\paragraph{Knowledge Retriever}
The retriever is defined using a dense inner product model:
\begin{align*}
p(z \mid x) &= \frac{\exp{f(x, z)}}{\sum_{z'} \exp{f(x,z')}}, \\
f(x, z) &= \inputembed(x)^\top \docembed(z),
\end{align*}
where $\inputembed$ and $\docembed$ are embedding
functions that map $x$ and $z$ respectively to $d$-dimensional vectors.
The \emph{relevance score} $f(x, z)$ between $x$ and $z$ is defined as the inner product of the vector embeddings.
The retrieval distribution is the softmax over all relevance scores.

We implement the embedding functions using BERT-style Transformers \cite{bert}. Following standard practices, we join spans of text by applying wordpiece tokenization, separating them with $\sep$ tokens, prefixing a $\cls$ token, and appending a final $\sep$ token.
\begin{align*}
    \joinbert(x) & = \cls x \sep \\
    \joinbert(x_1, x_2) & = \cls x_1 \sep x_2 \sep
\end{align*}
As in \citet{bert}, we pass this into a Transformer, which produces one vector for each token, including the vector corresponding to $\cls$ which is used as a ``pooled'' representation of the sequence (denoted $\bertsub{CLS}$). Finally, we perform a linear projection to reduce the dimensionality of the vector, denoted as a projection matrix $\mathbf{W}$:\begin{align*}
\inputembed(x) & = \mathbf{W_\mathtt{input}}\bertsub{CLS}(\joinbert(x)) \\
\docembed(z) & = \mathbf{W_\mathtt{doc}}\bertsub{CLS}(\joinbert(z_\text{title}, z_\text{body}))
\end{align*}
where $z_\text{title}$ is the document's title and $z_\text{body}$ is its body.
We let $\theta$ denote all parameters associated with the retriever, which include the Transformer and projection matrices.

\paragraph{Knowledge-Augmented Encoder}
Given an input $x$ and a retrieved document $z$, the knowledge-augmented encoder defines $p(y\mid z,x)$. We join $x$ and $z$ into a single sequence that we feed into a Transformer (distinct from the one used in the retriever). This allows us to perform rich cross-attention between $x$ and $z$ before predicting $y$. See Figure~\ref{fig:intro} for a concrete example.

At this stage, the architectures for pre-training and fine-tuning differ slightly. For the masked language model pre-training task, we must predict the original value of each \mask token in $x$. To do so, we use the same masked language modeling (MLM) loss as in \citet{bert}:
\begin{align*}
\centering
p(y\mid z,x) &= \prod_{j=1}^{J_x} p(y_j \mid z, x) \\
p(y_j \mid z, x) &\propto \exp \left(w_j^\top \bert_{\mathtt{MASK}(j)}(\joinbert(x, z_\text{body}))\right)
\end{align*}
where $\bert_{\mathtt{MASK}(j)}$ denotes the Transformer output vector corresponding to the $j^{th}$ masked token, $J_x$ is the total number of \mask tokens in $x$, and $w_j$ is a learned word embedding for token $y_j$.

For \openqa fine-tuning, we wish to produce the answer string $y$. Following previous reading comprehension work \cite{squad,bidaf,rasor,bidaf_plusplus}, we will assume that the answer $y$ can be found as a contiguous sequence of tokens in some document $z$.
Let $S(z, y)$ be the set of spans matching $y$ in $z$. Then we can define $p(y \mid z,x)$ as:
\begin{align*}
p(y \mid z, x)  &\propto \sum_{s \in S(z, y)} \exp\left(\mathtt{MLP}\left(\left[h_{\mathtt{START(s)}} ; h_{\mathtt{END(s)}}\right]\right)\right)\\
h_{\mathtt{START(s)}} &= \bertsub{START(s)}(\joinbert(x, z_\text{body})), \\
h_{\mathtt{END(s)}} &= \bertsub{END(s)}(\joinbert(x, z_\text{body})),
\end{align*}
where $\bertsub{START(s)}$ and $\bertsub{END(s)}$ denote the Transformer output vectors corresponding to the start and end tokens of span $s$, respectively, while $\mathtt{MLP}$ denotes a feed-forward neural network. We will let $\phi$ denote all parameters associated with the knowledge-augmented encoder.

\subsection{Training}
\label{sec:training}

For both pre-training and fine-tuning, we train by maximizing the log-likelihood $\log p(y\mid x)$ of the correct output $y$. Since both the knowledge retriever and knowledge-augmented encoder are differentiable neural networks, we can compute the gradient of $\log p(y\mid x)$ (defined in Equation~\ref{eqn:marginal}) with respect to the model parameters $\theta$ and $\phi$, and optimize using stochastic gradient descent.

The key computational challenge is that the marginal probability
$p(y\mid x) = \sum_{z \in \cZ} p(y\mid x,z)\,p(z \mid x)$
involves a summation over all documents $z$ in the knowledge corpus $\cZ$. We approximate this by instead summing over the top $k$ documents with highest probability under $p(z \mid x)$---this is reasonable if most documents have near zero probability.

Even with this approximation, we still need an efficient way to find the top $k$ documents. Note that the ordering of documents under $p(z \mid  x)$ is the same as under the relevance score $f(x, z) = \inputembed(x)^\top \docembed(z)$, which is an inner product. Thus, we can employ Maximum Inner Product Search (MIPS) algorithms to find the approximate top $k$ documents, using running time and storage space that scale sub-linearly with the number of documents \cite{mips_cone,mips_alsh,mips_binary}.

To employ MIPS, we must pre-compute $\docembed(z)$ for every $z \in \cZ$  and construct an efficient search index over these embeddings. However, this data structure will no longer be consistent with $p(z \mid x)$ if the parameters $\theta$ of $\docembed$ are later updated. Hence, the search index goes ``stale'' after every gradient update on $\theta$.

Our solution is to ``refresh'' the index by asynchronously re-embedding and re-indexing all documents every several hundred training steps. The MIPS index is slightly stale between refreshes, but note that it is {\em only} used to select the top $k$ documents. We recompute $p(z \mid x)$ and its gradient, using the fresh $\theta$, for these top $k$ documents after retrieving them.
In Section~\ref{sec:ablation}, we empirically demonstrate that this procedure results in stable optimization, provided that refreshes happen at a sufficiently frequent rate.

\paragraph{Implementing asynchronous MIPS refreshes}
We asynchronously refresh the MIPS index by running two jobs in parallel: a primary {\em trainer} job, which performs gradient updates on the parameters, and a secondary {\em index builder} job, which embeds and indexes the documents. As shown below, the trainer sends the index builder a snapshot of its parameters, $\theta'$. The trainer then continues to train while the index builder uses $\theta'$ to construct a new index in the background. As soon as the index builder is done, it sends the new index back to the trainer, and the process repeats.

{
\begin{figure}
\begin{center}
\includegraphics[width=.8\columnwidth]{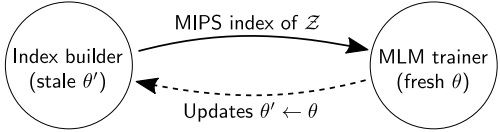}
\caption{\thename pre-training with asynchronous MIPS refreshes.}
\end{center}
\end{figure}
}

While asynchronous refreshes can be used for both pre-training and fine-tuning, in our experiments we only use it for pre-training. For fine-tuning, we just build the MIPS index once (using the pre-trained $\theta$) for simplicity and do not update $\docembed$.\footnote{This works because pre-training already yields a good $\docembed$ function. However, it is possible that refreshing the index would further improve performance.} Note that we still fine-tune $\inputembed$, so the retrieval function is still updated from the query side.

\paragraph{What does the retriever learn?}
Since the knowledge retrieval of \thename is latent, it is not obvious how the training objective encourages meaningful retrievals. Here, we show how it rewards retrievals that improve prediction accuracy.

For a given query $x$ and document $z$, recall that $f(x,z)$
is the ``relevance score'' that the knowledge retriever assigns to document $z$.
We can see how a single step of gradient descent during \thename pre-training alters this score by analyzing
the gradient with respect to the parameters of the knowledge retriever, $\theta$:
\begin{align*}
\nabla \log p(y \mid x) & =\sum_{z\in\cZ} r(z) \nabla f(x,z) \\
r(z) & =\left[\frac{p(y \mid z,x)}{p(y \mid x)}-1\right]p(z \mid x).
\end{align*}
For each document $z$, the gradient encourages
the retriever to change the score $f(x,z)$ by $r(z)$ --- 
increasing if $r(z)$ is positive, and decreasing if negative.
The multiplier $r(z)$
is positive if and only if $p(y\mid z,x) > p(y\mid x)$.
The term $p(y\mid z,x)$ is the probability of predicting
the correct output $y$ when using document $z$.
The term $p(y \mid x)$ is the expected value of $p(y \mid x, z)$ when randomly sampling a document from $p(z \mid x)$. Hence, document $z$ receives a positive update whenever it performs better than expected.

\subsection{Injecting inductive biases into pre-training} \label{sec:inductive-bias}
In the process of developing \thename, we discovered several additional strategies that further guide the model towards meaningful retrievals, described below.

\paragraph{Salient span masking}
During \thename pre-training, we want to focus on examples $x$ that require world knowledge to predict the masked tokens. As explained in Section~\ref{sec:background}, some MLM spans only require local context. To focus on problems that require world knowledge, we mask \emph{salient spans} such as \nl{United Kingdom} or \nl{July 1969}. We use a BERT-based tagger trained on CoNLL-2003 data \cite{conll_ner} to identify named entities, and a regular expression to identify dates. We select and mask one of these salient spans within a sentence for the masked language modeling task. We show that this significantly outperforms other masking strategies in Section~\ref{sec:ablation}.

\paragraph{Null document}
Even with salient span masking, not all masked tokens require world knowledge to predict. We model this by adding an empty \emph{null document} $\znull$~to the top $k$ retrieved documents, allowing appropriate credit to be assigned to a consistent sink when no retrieval is necessary.

\paragraph{Prohibiting trivial retrievals}
If the pre-training corpus $\cX$ and the knowledge corpus $\cZ$ are the same, there exists a trivial retrieval candidate $z$ that is \emph{too} informative: if the masked sentence $x$ comes from document $z$, the knowledge augmented encoder can trivially predict $y$ by looking at the unmasked version of $x$ in $z$. This results in a large positive gradient for $p(z\mid x)$. If this occurs too often, the knowledge retriever ends up learning to look for exact string matches between $x$ and $z$, which does not capture other forms of relevance. For this reason, we exclude this trivial candidate during pre-training.

\paragraph{Initialization}
At the beginning of training, if the retriever does not have good embeddings for $\inputembed(x)$ and $\docembed(z)$, the retrieved documents $z$ will likely be unrelated to $x$. This causes the knowledge augmented encoder to learn to ignore the retrieved documents. Once this occurs, the knowledge retriever does not receive a meaningful gradient and cannot improve, creating a vicious cycle. To avoid this cold-start problem, we warm-start $\inputembed$ and $\docembed$ using a simple training objective known as the Inverse Cloze Task (ICT) where, given a sentence, the model is trained to retrieve the document where that sentence came from. We defer to ~\citet{orqa} for details. For the knowledge-augmented encoder, we warm-start it with BERT pre-training---specifically, the uncased BERT-base model (12 layers, 768 hidden units, 12 attention heads).
 \section{Experiments} \label{sec:experiments}

We now evaluate our approach on the \openqa task.
In this section, we describe in detail the benchmarks used and the different approaches to which we compare empirically.

\subsection{\openqa Benchmarks}
A number of benchmarks have been proposed for \openqa. In this work, we focus on datasets where the question writers did not already know the answer. This yields questions that reflect more realistic information-seeking needs, and also avoids artifacts that can arise if the question is formulated with a particular answer in mind. A deeper justification is given in ~\citet{orqa}.
In all cases, the predicted answer is evaluated via exact match with any reference answer, following previous \openqa work~\cite{drqa}.

\paragraph{NaturalQuestions-Open} The NaturalQuestions dataset \cite{naturalquestions} consists of naturally occurring Google queries and their answers. Each answer also comes with an ``answer type'': following \citet{orqa}, we only keep questions that are categorized as ``short answer type'' with at most five tokens. The dataset also provides a suggested Wikipedia document to retrieve; like all models we compare against, we do not provide this to our model.

\paragraph{WebQuestions} The WebQuestions dataset \cite{webquestions} was collected from the Google Suggest API, using one seed question and expanding the set to related questions. We follow the setting defined by \citet{drqa}.

\paragraph{CuratedTrec}
The CuratedTrec dataset is a collection of question-answer pairs drawn from real user queries issued on sites such as MSNSearch and AskJeeves. To account for multiple correct answers or different spelling variations, the answers in this dataset are defined as regular expressions that match all correct answers. It is unclear how to train generation-based models with this type of supervision, so we do not evaluate them on this dataset.

\subsection{Approaches compared}
\label{sec:comparisons}
\paragraph{Retrieval-based \openqa}
Most existing \openqa systems answer the input question by first retrieving potentially relevant documents from a knowledge corpus, and then using a reading comprehension system to extract an answer from the documents. In this paradigm, the knowledge is stored \emph{explicitly} in the corpus.
We wish to compare different methods for implementing retrieval.

\hyphenation{HardEM}
Many approaches use non-learned heuristic retrieval such as sparse bag-of-words matching~\cite{bm25} or entity linking on the question to select a small set of relevant documents (e.g., 20). These documents are typically then re-ranked using a learned model, but coverage may be limited by the initial heuristic retrieval step. Approaches such as DrQA~\cite{drqa}, HardEM~\cite{openqa_hardem}, GraphRetriever~\cite{GraphRetriever}, and PathRetriever~\cite{rrp_salesforce} in Table~\ref{tab:main_results} are in this category.

Some recent approaches have proposed to implement learnable retrieval using a MIPS index. ORQA~\cite{orqa} formulates \openqa using a similar latent variable model as \thename, and also trains by maximizing the marginal likelihood. However, \thename adds a novel language model pre-training step, and backpropagates into the MIPS index, rather than using a fixed index. In Table~\ref{tab:main_results}, we directly compare the two. It is also important to note that the retrievers for both \thename pretraining and ORQA are initialized using the Inverse Cloze Task, described in Section~\ref{sec:inductive-bias}.

\paragraph{Generation-based \openqa}
An emerging alternative approach to \openqa is to model it as a sequence prediction task:
simply encode the question, and then decode the answer token-by-token based on the encoding.
While it was initially unclear how large amounts of knowledge could be injected into the model,
GPT-2~\cite{gpt2} hinted at the possibility of directly generating answers without using any given context via sequence-to-sequence. However, their performance was not competitive possibly due to the lack of fine-tuning. Orthogonally, T5~\cite{t5} showed that directly generating answers without explicit extraction from the given context is viable approach, but they only experimented on the reading comprehension task, where a context document is provided.

For the most competitive and comparable generation-based baseline, we compare to concurrent work which fine-tunes T5 for \openqa~\cite{t5_openqa}.\footnote{We initially conducted our own T5 experiments using the code from {\tiny\url{https://tinyurl.com/t5-openqa-colab}}~\cite{t5}. We now report results from the concurrent work of \citet{t5_openqa}, which has an improved fine-tuning procedure.}
We compare against the Base, Large, and even larger 11-billion parameter model to measure the effect of model size.

\begin{table*}[t!]
\caption{Test results on \openqa benchmarks. The number of train/test examples are shown in paretheses below each benchmark. Predictions are evaluated with exact match against any reference answer. Sparse retrieval denotes methods that use sparse features such as TF-IDF and BM25. Our model, \thename, outperforms all existing systems. }
\vspace{.1in}
\centering
\footnotesize
\begin{tabular}{@{}lllcccr@{}}
\toprule
\textbf{Name} & \textbf{Architectures}
& \makecell[l]{\textbf{Pre-training}}
& \makecell{\textbf{NQ}\\(79k/4k)} & \makecell{\textbf{WQ}\\(3k/2k)} & \makecell{\textbf{CT}\\(1k /1k)} & \textbf{\# params}\\
\midrule
 BERT-Baseline~\cite{orqa} & Sparse Retr.$+$Transformer & BERT& 26.5 & 17.7 & 21.3 & 110m\\
\cmidrule{1-7}
T5 (base)~\cite{t5_openqa} & Transformer Seq2Seq & T5 (Multitask) & 27.0 & 29.1  & -  & 223m\\
T5 (large)~\cite{t5_openqa} & Transformer Seq2Seq & T5 (Multitask)& 29.8 & 32.2  & -  & 738m\\
T5 (11b)~\cite{t5_openqa} & Transformer Seq2Seq & T5 (Multitask) & 34.5 & 37.4 & -  & 11318m\\
\cmidrule{1-7}
DrQA~\cite{drqa} & Sparse Retr.$+$DocReader & N/A& - & 20.7 & 25.7   & 34m\\
HardEM~\cite{openqa_hardem} &Sparse Retr.$+$Transformer & BERT& 28.1 & - & - & 110m
\\
GraphRetriever~\cite{GraphRetriever} & GraphRetriever$+$Transformer& BERT & 31.8 & 31.6 & - & 110m
\\
PathRetriever~\cite{rrp_salesforce} & PathRetriever$+$Transformer&MLM&  32.6 & - & -  & 110m\\
ORQA~\cite{orqa} & Dense Retr.$+$Transformer & ICT$+$BERT & 33.3 & 36.4 & 30.1  & 330m\\
\cmidrule{1-7}
Ours (\target= Wikipedia, \unlabeled= Wikipedia) & Dense Retr.$+$Transformer & \thename & 39.2 & 40.2 & \textbf{46.8} & 330m\\
Ours (\target= CC-News, \unlabeled= Wikipedia) & Dense Retr.$+$Transformer & \thename & \textbf{40.4} & \textbf{40.7} & 42.9 & 330m\\
\bottomrule
\vspace{.02in}
\end{tabular}
\label{tab:main_results}
\end{table*}

\subsection{Implementation Details}
\paragraph{Fine-tuning}
We reuse all hyperparameters from \citet{orqa}, to enable direct comparison. Our knowledge corpus is derived from the December 20, 2018 snapshot of English Wikipedia. Documents are greedily split into chunks of up to 288 BERT wordpieces, resulting in just over 13 million retrieval candidates. During fine-tuning inference, we consider the top-5 candidates, and the entire model can be run on a single machine with a 12GB GPU.

\paragraph{Pre-training}
We pre-train for 200k steps on 64 Google Cloud TPUs, with a batch size of 512 and a learning rate of 3e-5, using BERT's default optimizer. The document embedding step for the MIPS index is parallelized over 16 TPUs. For each example, we retrieve and marginalize over 8 candidate documents, including the null document $\znull$.

We experiment with two choices of the pre-training corpus \target:
(1)~Wikipedia, which is identical to the knowledge corpus $\cZ$, and (2)~CC-News, our reproduction of the corpus of English news proposed by~\citet{roberta}.

{
\begin{table}[t!]
\caption{Ablation experiments on NQ's development set.}
\vspace{.1in}
\footnotesize \centering
\begin{tabular}{@{}lcc@{}}
\toprule
\textbf{Ablation} & \makecell{\textbf{Exact}\\\textbf{Match}} & \makecell{\textbf{Zero-shot}\\\textbf{Retrieval}\\\textbf{Recall@5}}\\
\midrule
\thename & 38.2 & 38.5 \\
\cmidrule{1-3}
REALM retriever$+$Baseline encoder  & 37.4 & 38.5\\
Baseline retriever$+$REALM encoder & 35.3 & 13.9 \\
Baseline (ORQA) & 31.3 & 13.9\\
\cmidrule{1-3}
\thename with random uniform masks & 32.3 & 24.2\\
\thename with random span masks & 35.3 & 26.1\\
\cmidrule{1-3}
30$\times$ stale MIPS & 28.7 & 15.1 \\
\bottomrule
\end{tabular}
\label{tab:ablations}
\end{table}
} \begin{table*}[t!]
\caption{An example where REALM utilizes retrieved documents to better predict masked tokens. It assigns much higher probability (0.129) to the correct term, \nl{Fermat}, compared to BERT. (Note that the blank corresponds to 3 BERT wordpieces.)}
\vspace{.1in}
\footnotesize \centering
\begin{tabular}{@{}rl@{\;\;}l@{\;\;}ll@{}} \toprule
$x$:
& \multicolumn{4}{l}{{An equilateral triangle is easily constructed using a straightedge and compass, because 3 is a \blank~prime.}}
\\
\midrule
(a)~~~~~BERT & $p(y=\text{\nl{Fermat}}\mid x)$ &$=$& $1.1\times10^{-14}$ & (No retrieval.) \\
\midrule
(b)~\thename& $p(y=\text{\nl{Fermat}}\mid x, z)$ &$=$& $1.0$
& (Conditional probability with document $z=$``257 is~\dots~a Fermat prime. \\
& & & & Thus a regular polygon with 257 sides is constructible with compass \dots'') \\
(c)~\thename& $p(y=\text{\nl{Fermat}}\mid x)$ &$=$& $0.129$ & (Marginal probability, marginalizing over top 8 retrieved documents.) \\ 
\bottomrule
\end{tabular}
\label{tab:retrieval_examples}
\end{table*} 
\subsection{Main results}
Table~\ref{tab:main_results} shows the accuracy of different approaches on the three \openqa datasets. \thename outperform all previous approaches by a significant margin. Table~\ref{tab:main_results} also shows the number of parameters for each model. 

As reported in the concurrent work of~\citet{t5_openqa}, the generative \openqa systems based on T5 are surprisingly powerful, with the largest T5-11B model outperforming the previous best \openqa system.
Increasing the size of T5 yields consistent improvement, but comes at significant computational cost (from Base to 11B, the model is 50 times larger, and gains roughly 5 points in accuracy). In contrast, \thename outperforms the largest T5-11B model while being 30 times smaller. It is also important to note that T5 accesses additional reading comprehension data from SQuAD during its pre-training (100,000+ examples). Access to such data could also benefit \thename, but was not used in our experiments.

Among all systems, the most direct comparison with \thename is ORQA~\cite{orqa}, where the fine-tuning setup, hyperparameters and training data are identical. The improvement of \thename over ORQA is purely due to better pre-training methods. The results also indicate that our method of pre-training can be applied both on (1) the single-corpus setting (\target= Wikipedia, \unlabeled= Wikipedia), or (2) the separate-corpus setting (\target= CC-News, \unlabeled= Wikipedia).

Compared to other retrieval-based systems~\cite{rrp_salesforce,openqa_hardem,GraphRetriever} which often retrieve from 20 to 80 documents, our system gets the overall best performance while only retrieving 5 documents.

\subsection{Analysis}\label{sec:ablation}

In Table~\ref{tab:ablations} we present results for NaturalQuestions-Open after ablating critical components of \thename. In addition to the end-to-end results, we also report how often the gold answer appears in the top-5 retrievals before applying any fine-tuning. The latter metric more significantly isolates the contribution of improving the retriever during pre-training.

\paragraph{Encoder or Retriever}
We first aim to determine whether \thename pre-training improves the retriever or the encoder, or both. To do so, we can reset the parameters of either the retriever or the encoder to their baseline state before \thename pre-training, and feed that into fine-tuning. Resetting both the retriever and encoder reduces the system to our main baseline, ORQA. We find that both the encoder and retriever benefit from \thename training separately, but the best result requires both components acting in unison.

\paragraph{Masking scheme}
We compare our salient span masking scheme (Section~\ref{sec:inductive-bias})
with (1) random token masking introduced in BERT~\cite{bert}
and (2) random span masking proposed by SpanBERT~\cite{spanbert}. 
While such salient span masking has not been shown to be impactful in previous work with standard BERT training~\cite{spanbert}, it is crucial for \thename. Intuitively, the latent variable learning relies heavily on the utility of retrieval and is therefore more sensitive to a consistent learning signal.

\paragraph{MIPS index refresh rate}
During pre-training, we run a parallel process to re-embed corpus documents and rebuild the MIPS index.
This results in one index refresh per approximately 500 training steps.
To demonstrate the importance of frequent index refreshes,
we compare against using a slower refresh rate.
The results in Table~\ref{tab:ablations} suggests that a stale index can hurt model training, and further reducing this staleness could offer better optimization.

\paragraph{Examples of retrieved documents}
Table~\ref{tab:retrieval_examples} shows an example of
the \thename masked language model prediction.
In this example, \nl{Fermat}
is the correct word, and \thename (row~(c)) gives the word a much high probability
compared to the BERT model (row~(a)).
Since \thename manages to retrieve some documents with a related fact (row~(b)), the marginalized probability of the correct answer dramatically increases. This shows that \thename
is able to retrieve document to fill in the masked word even though it is trained with unsupervised text only.

\section{Discussion and Related Work}
We previously discussed related methods for \openqa.
Here we present several alternate ways of viewing REALM that connect it to a broader set of ideas beyond \openqa:

\paragraph{Language modeling with corpus as context}
Language representation models have been incorporating contexts of 
increasingly large scope when making predictions.
Examples of this progression include models that condition on surrounding words~\cite{skipgram, word2vec}, sentences~\cite{skipthought, elmo}, and paragraphs~\cite{gpt, bert}. We can view \thename as a generalization of the above work to the next level of scope: the entire text \emph{corpus}. 

\paragraph{Retrieve-and-edit with learned retrieval}
In order to better explain the variance in the input text and enable controllable generation, \citet{prototypes} proposed a language model with the retrieve-and-edit framework~\cite{retrieve_and_edit} that conditions on text with high lexical overlap. \thename has a similar approach, except that the model learns for itself which texts are most useful for reducing perplexity. By jointly learning the retriever, \thename has the capacity to depend on information beyond lexical overlap.

\paragraph{Scalable grounded neural memory}
The document index can be viewed as a memory where the keys are the document embeddings.
From this view, our work share motivations
with works such as product key memory~\cite{product_key},
which enables
sub-linear memory access in a memory network \cite{memory_networks,neural_turing,end_to_end_memory}, allowing these scalable memory layers to be integrated into large language models.
One main difference is that our memories are grounded---each memory is associated with a document rather than unnamed value vectors. This level of interpretability is crucial for applications like \openqa, where users would require provenance for a predicted answer to be trustworthy.

\paragraph{Unsupervised Corpus Alignment}
In sequence-to-sequence models with attention~\cite{attention}, text is generated with latent selection of relevant tokens. This results in a set of \emph{model-centric} unsupervised alignments between target and source tokens. Analogously, \thename also generates text with latent selection of relevant documents. A by-product of our method is that we offer a set of \emph{model-centric} unsupervised alignments between text in the pre-training corpus $\mathcal{X}$ and knowledge corpus $\mathcal{Z}$.

 \section{Future Work}
The work presented here is the minimal instantiation of a family of \thename-like approaches where a representation is pre-trained to perform reasoning over a large corpus of knowledge on-the-fly during inference. We are particularly optimistic about generalizations of this work to (1) structured knowledge, which would result in a generalization of \citet{knowbert} where we would also learn the decision of which entities are informative, (2) the multi-lingual setting, e.g., retrieving knowledge in a high-resource language to better represent text in a low-resource language,  and (3) the multi-modal setting, e.g., retrieving images or videos that can provide knowledge rarely observed in text. 
\bibliography{main}

\begin{thebibliography}{42}
\providecommand{\natexlab}[1]{#1}
\providecommand{\url}[1]{\texttt{#1}}
\expandafter\ifx\csname urlstyle\endcsname\relax
  \providecommand{\doi}[1]{doi: #1}\else
  \providecommand{\doi}{doi: \begingroup \urlstyle{rm}\Url}\fi

\bibitem[Asai et~al.(2019)Asai, Hashimoto, Hajishirzi, Socher, and
  Xiong]{rrp_salesforce}
Asai, A., Hashimoto, K., Hajishirzi, H., Socher, R., and Xiong, C.
\newblock Learning to retrieve reasoning paths over wikipedia graph for
  question answering.
\newblock \emph{arXiv preprint arXiv:1911.10470}, 2019.

\bibitem[Bahdanau et~al.(2014)Bahdanau, Cho, and Bengio]{attention}
Bahdanau, D., Cho, K., and Bengio, Y.
\newblock Neural machine translation by jointly learning to align and
  translate.
\newblock \emph{arXiv preprint arXiv:1409.0473}, 2014.

\bibitem[Berant et~al.(2013)Berant, Chou, Frostig, and Liang]{webquestions}
Berant, J., Chou, A., Frostig, R., and Liang, P.
\newblock Semantic parsing on freebase from question-answer pairs.
\newblock In \emph{Proceedings of the 2013 Conference on Empirical Methods in
  Natural Language Processing}, pp.\  1533--1544, 2013.

\bibitem[Brill et~al.(2002)Brill, Dumais, and Banko]{askmsr}
Brill, E., Dumais, S., and Banko, M.
\newblock An analysis of the askmsr question-answering system.
\newblock In \emph{Empirical Methods in Natural Language Processing}, 2002.

\bibitem[Chen et~al.(2017)Chen, Fisch, Weston, and Bordes]{drqa}
Chen, D., Fisch, A., Weston, J., and Bordes, A.
\newblock Reading wikipedia to answer open-domain questions.
\newblock In \emph{Proceedings of the 55th Annual Meeting of the Association
  for Computational Linguistics (Volume 1: Long Papers)}, volume~1, pp.\
  1870--1879, 2017.

\bibitem[Clark \& Gardner(2017)Clark and Gardner]{bidaf_plusplus}
Clark, C. and Gardner, M.
\newblock Simple and effective multi-paragraph reading comprehension.
\newblock In \emph{Annual Meeting of the Association for Computational
  Linguistics}, 2017.

\bibitem[Dai \& Le(2015)Dai and Le]{dai_finetune}
Dai, A.~M. and Le, Q.~V.
\newblock Semi-supervised sequence learning.
\newblock In \emph{Advances in neural information processing systems}, pp.\
  3079--3087, 2015.

\bibitem[Devlin et~al.(2018)Devlin, Chang, Lee, and Toutanova]{bert}
Devlin, J., Chang, M.-W., Lee, K., and Toutanova, K.
\newblock Bert: Pre-training of deep bidirectional transformers for language
  understanding.
\newblock \emph{arXiv preprint arXiv:1810.04805}, 2018.

\bibitem[Graves et~al.(2014)Graves, Wayne, and Danihelka]{neural_turing}
Graves, A., Wayne, G., and Danihelka, I.
\newblock Neural turing machines.
\newblock \emph{ArXiv}, abs/1410.5401, 2014.

\bibitem[Guu et~al.(2018)Guu, Hashimoto, Oren, and Liang]{prototypes}
Guu, K., Hashimoto, T.~B., Oren, Y., and Liang, P.
\newblock Generating sentences by editing prototypes.
\newblock \emph{Transactions of the Association for Computational Linguistics},
  6:\penalty0 437--450, 2018.

\bibitem[Hashimoto et~al.(2018)Hashimoto, Guu, Oren, and
  Liang]{retrieve_and_edit}
Hashimoto, T.~B., Guu, K., Oren, Y., and Liang, P.~S.
\newblock A retrieve-and-edit framework for predicting structured outputs.
\newblock In \emph{Advances in Neural Information Processing Systems}, pp.\
  10052--10062, 2018.

\bibitem[Joshi et~al.(2019)Joshi, Chen, Liu, Weld, Zettlemoyer, and
  Levy]{spanbert}
Joshi, M., Chen, D., Liu, Y., Weld, D.~S., Zettlemoyer, L., and Levy, O.
\newblock {SpanBERT}: Improving pre-training by representing and predicting
  spans.
\newblock \emph{arXiv preprint arXiv:1907.10529}, 2019.

\bibitem[Khandelwal et~al.(2019)Khandelwal, Levy, Jurafsky, Zettlemoyer, and
  Lewis]{knnlm}
Khandelwal, U., Levy, O., Jurafsky, D., Zettlemoyer, L., and Lewis, M.
\newblock Generalization through memorization: Nearest neighbor language
  models.
\newblock \emph{ArXiv}, abs/1911.00172, 2019.

\bibitem[Kiros et~al.(2015)Kiros, Zhu, Salakhutdinov, Zemel, Urtasun, Torralba,
  and Fidler]{skipthought}
Kiros, R., Zhu, Y., Salakhutdinov, R.~R., Zemel, R., Urtasun, R., Torralba, A.,
  and Fidler, S.
\newblock Skip-thought vectors.
\newblock In \emph{Advances in neural information processing systems}, pp.\
  3294--3302, 2015.

\bibitem[Kwiatkowski et~al.(2019)Kwiatkowski, Palomaki, Rhinehart, Collins,
  Parikh, Alberti, Epstein, Polosukhin, Kelcey, Devlin,
  et~al.]{naturalquestions}
Kwiatkowski, T., Palomaki, J., Rhinehart, O., Collins, M., Parikh, A., Alberti,
  C., Epstein, D., Polosukhin, I., Kelcey, M., Devlin, J., et~al.
\newblock Natural questions: a benchmark for question answering research.
\newblock \emph{Transactions of the Association for Computational Linguistics},
  2019.

\bibitem[Lample et~al.(2019)Lample, Sablayrolles, Ranzato, Denoyer, and
  J{\'e}gou]{product_key}
Lample, G., Sablayrolles, A., Ranzato, M., Denoyer, L., and J{\'e}gou, H.
\newblock Large memory layers with product keys.
\newblock In \emph{Advances in Neural Information Processing Systems}, pp.\
  8546--8557, 2019.

\bibitem[Lee et~al.(2016)Lee, Salant, Kwiatkowski, Parikh, Das, and
  Berant]{rasor}
Lee, K., Salant, S., Kwiatkowski, T., Parikh, A., Das, D., and Berant, J.
\newblock Learning recurrent span representations for extractive question
  answering.
\newblock \emph{arXiv preprint arXiv:1611.01436}, 2016.

\bibitem[Lee et~al.(2019)Lee, Chang, and Toutanova]{orqa}
Lee, K., Chang, M.-W., and Toutanova, K.
\newblock Latent retrieval for weakly supervised open domain question
  answering.
\newblock In \emph{Proceedings of the Conference of Association for
  Computational Linguistics}, 2019.

\bibitem[Lewis et~al.(2019)Lewis, Liu, Goyal, Ghazvininejad, Mohamed, Levy,
  Stoyanov, and Zettlemoyer]{bart_not_bert}
Lewis, M., Liu, Y., Goyal, N., Ghazvininejad, M., Mohamed, A., Levy, O.,
  Stoyanov, V., and Zettlemoyer, L.
\newblock Bart: Denoising sequence-to-sequence pre-training for natural
  language generation, translation, and comprehension.
\newblock \emph{ArXiv}, abs/1910.13461, 2019.

\bibitem[Liu et~al.(2019)Liu, Ott, Goyal, Du, Joshi, Chen, Levy, Lewis,
  Zettlemoyer, and Stoyanov]{roberta}
Liu, Y., Ott, M., Goyal, N., Du, J., Joshi, M., Chen, D., Levy, O., Lewis, M.,
  Zettlemoyer, L., and Stoyanov, V.
\newblock Roberta: A robustly optimized bert pretraining approach.
\newblock \emph{arXiv preprint arXiv:1907.11692}, 2019.

\bibitem[Mikolov et~al.(2013{\natexlab{a}})Mikolov, Chen, Corrado, and
  Dean]{skipgram}
Mikolov, T., Chen, K., Corrado, G., and Dean, J.
\newblock Efficient estimation of word representations in vector space.
\newblock \emph{arXiv preprint arXiv:1301.3781}, 2013{\natexlab{a}}.

\bibitem[Mikolov et~al.(2013{\natexlab{b}})Mikolov, Sutskever, Chen, Corrado,
  and Dean]{word2vec}
Mikolov, T., Sutskever, I., Chen, K., Corrado, G.~S., and Dean, J.
\newblock Distributed representations of words and phrases and their
  compositionality.
\newblock In \emph{Advances in neural information processing systems}, pp.\
  3111--3119, 2013{\natexlab{b}}.

\bibitem[Miller et~al.(2016)Miller, Fisch, Dodge, Karimi, Bordes, and
  Weston]{key_value_memorynetwork}
Miller, A., Fisch, A., Dodge, J., Karimi, A.-H., Bordes, A., and Weston, J.
\newblock Key-value memory networks for directly reading documents.
\newblock \emph{arXiv preprint arXiv:1606.03126}, 2016.

\bibitem[Min et~al.(2019{\natexlab{a}})Min, Chen, Hajishirzi, and
  Zettlemoyer]{openqa_hardem}
Min, S., Chen, D., Hajishirzi, H., and Zettlemoyer, L.
\newblock A discrete hard em approach for weakly supervised question answering.
\newblock \emph{arXiv preprint arXiv:1909.04849}, 2019{\natexlab{a}}.

\bibitem[Min et~al.(2019{\natexlab{b}})Min, Chen, Zettlemoyer, and
  Hajishirzi]{GraphRetriever}
Min, S., Chen, D., Zettlemoyer, L., and Hajishirzi, H.
\newblock Knowledge guided text retrieval and reading for open domain question
  answering.
\newblock \emph{arXiv preprint arXiv:1911.03868}, 2019{\natexlab{b}}.

\bibitem[Peters et~al.(2018)Peters, Neumann, Iyyer, Gardner, Clark, Lee, and
  Zettlemoyer]{elmo}
Peters, M.~E., Neumann, M., Iyyer, M., Gardner, M., Clark, C., Lee, K., and
  Zettlemoyer, L.
\newblock Deep contextualized word representations.
\newblock In \emph{Proc. of NAACL}, 2018.

\bibitem[Peters et~al.(2019)Peters, Neumann, IV, Schwartz, Joshi, Singh, and
  Smith]{knowbert}
Peters, M.~E., Neumann, M., IV, R. L.~L., Schwartz, R., Joshi, V., Singh, S.,
  and Smith, N.~A.
\newblock Knowledge enhanced contextual word representations, 2019.

\bibitem[Petroni et~al.(2019)Petroni, Rockt{\"a}schel, Lewis, Bakhtin, Wu,
  Miller, and Riedel]{lm_as_kb}
Petroni, F., Rockt{\"a}schel, T., Lewis, P., Bakhtin, A., Wu, Y., Miller,
  A.~H., and Riedel, S.
\newblock Language models as knowledge bases?
\newblock \emph{arXiv preprint arXiv:1909.01066}, 2019.

\bibitem[Radford et~al.(2018)Radford, Narasimhan, Salimans, and Sutskever]{gpt}
Radford, A., Narasimhan, K., Salimans, T., and Sutskever, I.
\newblock Improving language understanding with unsupervised learning.
\newblock Technical report, OpenAI, 2018.

\bibitem[Radford et~al.(2019)Radford, Wu, Child, Luan, Amodei, and
  Sutskever]{gpt2}
Radford, A., Wu, J., Child, R., Luan, D., Amodei, D., and Sutskever, I.
\newblock Language models are unsupervised multitask learners.
\newblock \emph{OpenAI Blog}, 2019.

\bibitem[Raffel et~al.(2019)Raffel, Shazeer, Roberts, Lee, Narang, Matena,
  Zhou, Li, and Liu]{t5}
Raffel, C., Shazeer, N., Roberts, A., Lee, K., Narang, S., Matena, M., Zhou,
  Y., Li, W., and Liu, P.~J.
\newblock Exploring the limits of transfer learning with a unified text-to-text
  transformer.
\newblock \emph{arXiv preprint arXiv:1910.10683}, 2019.

\bibitem[Rajpurkar et~al.(2016)Rajpurkar, Zhang, Lopyrev, and Liang]{squad}
Rajpurkar, P., Zhang, J., Lopyrev, K., and Liang, P.
\newblock Squad: 100,000+ questions for machine comprehension of text.
\newblock In \emph{Proceedings of the 2016 Conference on Empirical Methods in
  Natural Language Processing}, pp.\  2383--2392, 2016.

\bibitem[Rajpurkar et~al.(2018)Rajpurkar, Jia, and Liang]{squad2}
Rajpurkar, P., Jia, R., and Liang, P.
\newblock Know what you don't know: Unanswerable questions for squad.
\newblock \emph{arXiv preprint arXiv:1806.03822}, 2018.

\bibitem[Ram \& Gray(2012)Ram and Gray]{mips_cone}
Ram, P. and Gray, A.~G.
\newblock Maximum inner-product search using cone trees.
\newblock In \emph{Proceedings of the 18th ACM SIGKDD international conference
  on Knowledge discovery and data mining}, pp.\  931--939, 2012.

\bibitem[Roberts et~al.(2020)Roberts, Raffel, and Shazeer]{t5_openqa}
Roberts, A., Raffel, C., and Shazeer, N.
\newblock How much knowledge can you pack into the parameters of a language
  model?
\newblock \emph{arXiv preprint arXiv:TBD}, 2020.

\bibitem[Robertson et~al.(2009)Robertson, Zaragoza, et~al.]{bm25}
Robertson, S., Zaragoza, H., et~al.
\newblock The probabilistic relevance framework: Bm25 and beyond.
\newblock \emph{Foundations and Trends in Information Retrieval}, 3\penalty0
  (4):\penalty0 333--389, 2009.

\bibitem[Sang \& De~Meulder(2003)Sang and De~Meulder]{conll_ner}
Sang, E. T.~K. and De~Meulder, F.
\newblock Introduction to the conll-2003 shared task: Language-independent
  named entity recognition.
\newblock In \emph{Proceedings of the Seventh Conference on Natural Language
  Learning at HLT-NAACL 2003}, pp.\  142--147, 2003.

\bibitem[Seo et~al.(2016)Seo, Kembhavi, Farhadi, and Hajishirzi]{bidaf}
Seo, M., Kembhavi, A., Farhadi, A., and Hajishirzi, H.
\newblock Bidirectional attention flow for machine comprehension.
\newblock In \emph{International Conference on Learning Representations}, 2016.

\bibitem[Shen et~al.(2015)Shen, Liu, Zhang, Yang, and Tao~Shen]{mips_binary}
Shen, F., Liu, W., Zhang, S., Yang, Y., and Tao~Shen, H.
\newblock Learning binary codes for maximum inner product search.
\newblock In \emph{Proceedings of the IEEE International Conference on Computer
  Vision}, pp.\  4148--4156, 2015.

\bibitem[Shrivastava \& Li(2014)Shrivastava and Li]{mips_alsh}
Shrivastava, A. and Li, P.
\newblock Asymmetric lsh (alsh) for sublinear time maximum inner product search
  (mips).
\newblock In \emph{Advances in Neural Information Processing Systems}, pp.\
  2321--2329, 2014.

\bibitem[Sukhbaatar et~al.(2015)Sukhbaatar, Weston, Fergus,
  et~al.]{end_to_end_memory}
Sukhbaatar, S., Weston, J., Fergus, R., et~al.
\newblock End-to-end memory networks.
\newblock In \emph{Advances in neural information processing systems}, 2015.

\bibitem[Weston et~al.(2014)Weston, Chopra, and Bordes]{memory_networks}
Weston, J., Chopra, S., and Bordes, A.
\newblock Memory networks.
\newblock \emph{arXiv preprint arXiv:1410.3916}, 2014.

\end{thebibliography}
\vfill
\pagebreak
\appendix
\begin{table*}[ptb]
\centering
\begin{tabular}{@{}l@{ }rl@{}} \toprule
$x$: & \multicolumn{2}{l}{\nl{Jennifer \blank~formed the production company Excellent Cadaver.}} \\ \midrule
BERT & &
also (0.13), then (0.08), later (0.05), \dots \\
REALM & (\unlabeled=20 Dec 2018 corpus) &
smith (0.01), brown (0.01), jones (0.01) \\
REALM & (\unlabeled=20 Jan 2020 corpus) & 
\textbf{lawrence} (0.13), brown (0.01), smith (0.01), \dots \\
\bottomrule
\end{tabular}

\caption{An example where REALM adapts to the updated knowledge corpus. The 
Wikipedia page ``Excellent Cadaver'' was added in 2019, so the model was not about to recover
the word when the knowledge corpus is outdated (2018). Interestingly, the same REALM model pre-trained on the 2018 corpus is able to retrieve the document in the updated corpus (2020) and generate the correct token, \nl {Lawrence}.}
\label{tab:adapting}
\end{table*} 
\section{Derivation of the gradient with respect to the knowledge retriever}

We compute the gradient of the \thename pre-training objective (a log-likelihood)
with respect to the parameters of the knowledge retriever, $\theta$:
\begin{align*}
\nabla\log p(y\mid x) &= p(y\mid x)^{-1}\nabla p(y\mid x) \\
 &= p(y\mid x)^{-1}\sum_{z}p(y\mid z,x)\nabla p(z\mid x) \\
 &= p(y\mid x)^{-1}\sum_{z}p(y\mid z,x)p(z\mid x)\nabla\log p(z\mid x) \\
 & =\sum_{z}p(z\mid y,x)\nabla\log p(z\mid x),
\end{align*}
where the last line follows from applying conditional Bayes' rule.
We can then expand $\nabla\log p\left(z\mid x\right)$ as:
\begin{align*}
\nabla\log p(z\mid x) &= \nabla\log\frac{\exp f(x,z)}{\sum_{z'}\exp f(x,z')} \\
 &= \nabla\left[f(x,z)-\log\sum_{z'}\exp f(x,z')\right] \\
 &= \nabla f(x,z)-\sum_{z'}p(z'\mid x)\nabla f(x,z')
\end{align*}
Plugging this back into the first set of equations yields:
\resizebox{0.48\textwidth}{!}{
\begin{minipage}{1.2\linewidth}
\begin{align*}
\nabla\log p\left(y\mid x\right) & =\sum_{z}p\left(z\mid y,x\right)\left[\nabla f(x,z)-\sum_{z'}p\left(z'\mid x\right)\nabla f(x,z')\right]\\
 & =\sum_{z}p\left(z\mid y,x\right)\nabla f(x,z)-\sum_{z'}p\left(z'\mid x\right)\nabla f(x,z')\\
 & =\sum_{z}\left[p\left(z\mid y,x\right)-p\left(z\mid x\right)\right]\nabla f(x,z)\\
 & =\sum_{z}\left[\frac{p\left(y\mid z,x\right)p\left(z\mid x\right)}{p\left(y\mid x\right)}-p\left(z\mid x\right)\right]\nabla f(x,z)\\
 & =\sum_{z}\left[\frac{p\left(y\mid z,x\right)}{p\left(y\mid x\right)}-1\right]p\left(z\mid x\right)\nabla f(x,z).
\end{align*}
\end{minipage}
}

In the second line, we used the fact that the overall expression is
an expectation with respect to $p\left(z\mid y,x\right)$, and the
terms which depend on $z'$ but not $z$ can be moved out of that
expectation.

\section{Connection between \thename and supervised learning}
From the equations in Appendix A, we saw that 
\[
\nabla\log p\left(y\mid x\right)=\sum_{z}\left[p\left(z\mid y,x\right)-p\left(z\mid x\right)\right]\nabla f(x,z).
\]
Suppose that there exists one document $z^{*}$ which causes the model
to achieve perfect prediction accuracy (i.e., $p\left(y\mid z^{*},x\right)=1$), while all other documents $z'$ result in zero accuracy (i.e., $p\left(y\mid z',x\right)=0$).
Under this setting, $p\left(z^{*}\mid y,x\right)=1$ (provided that
$p\left(z^{*}\mid x\right)$ is non-zero), which causes the gradient
to become
\begin{align*}
\nabla\log p\left(y\mid x\right) & =\nabla f\left(x,z^{*}\right)-\sum_{z}p\left(z\mid x\right)\nabla f(x,z)\\
 & =\nabla\log p\left(z^{*}\mid x\right).
\end{align*}
From this, we see that gradient descent on the \thename objective is
equivalent to gradient descent on $\log p\left(z^{*}\mid x\right)$.
This is none other than the typical maximum likelihood training objective
used in supervised learning, where $z^{*}$ is the ``gold'' document.

\section{Adapting to new knowledge}

An explicit retrieval system allows us to adapt to new world knowledge
simply by modifying the corpus documents.
To demonstrate this ability,
we replace the knowledge corpus with a more recent version of Wikipedia corpus after pre-training is done.
When the input query is about a fact where the two corpora disagree,
\thename can change the prediction to reflect the updated information,
as exemplified in Table~\ref{tab:adapting}.
However, even with an explicit retrieval mechanism,
the knowledge-augmented encoder will still end up remembering some world knowledge,
making the prediction of some input sentences not updated with the new corpus.
(For instance, the model predicts \nl{Thatcher} for \nl{\blank~is the prime minister of United Kingdom.} on both corpora, perhaps due to the frequent mention of her name in Wikipedia articles.)

\section{Retrieval Utility}
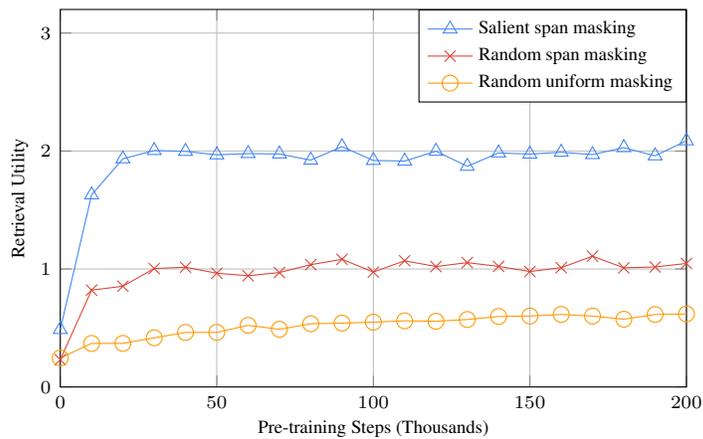
\begin{figure*}[ptb]
\centering
\begin{tikzpicture}
 \begin{axis}[
   width=1.2\columnwidth,
   height=0.8\columnwidth,
   legend cell align=left,
   legend style={at={(1, 1)},anchor=north east,font=\scriptsize},
   mark options={mark size=3},
   font=\scriptsize,
   xmin=0, xmax=200,
   ymin=0, ymax=3.2,
   xtick={0, 50, 100, 150, 200},
   ymajorgrids=true,
   xmajorgrids=true,
   xlabel style={yshift=0.5ex,},
   xlabel=Pre-training Steps (Thousands) ,
   ylabel=Retrieval Utility,
   ylabel style={yshift=-0.5ex,}]
    \addplot[mark=triangle,g-blue] plot coordinates {
(0, 0.4870335553570302)
(10, 1.6283367644777194)
(20, 1.933776036871063)
(30, 2.0041376407247413)
(40, 1.9967591197965078)
(50, 1.967421851377687)
(60, 1.9783062606785702)
(70, 1.9747018990660743)
(80, 1.9228911003560645)
(90, 2.036878828395492)
(100, 1.9200309391254438)
(110, 1.914781704899911)
(120, 1.9980497854538093)
(130, 1.8706547102603057)
(140, 1.9837496104468106)
(150, 1.973855775378632)
(160, 1.9895911115697824)
(170, 1.9696919231070622)
(180, 2.028405946627168)
(190, 1.9590725785261562)
(200, 2.086580118497257)
    };
    \addlegendentry{Salient span masking}
    \addplot[mark=x,g-red] plot coordinates {
(0, 0.23141121073211263)
(10, 0.8205503256147296)
(20, 0.854025527215577)
(30, 1.003707903515242)
(40, 1.0150897896138955)
(50, 0.9640269530993363)
(60, 0.9427491107832932)
(70, 0.9697737769948522)
(80, 1.0363693585282714)
(90, 1.0821630555977242)
(100, 0.9739717430562631)
(110, 1.0695464290750996)
(120, 1.02078362435653)
(130, 1.0538118199206679)
(140, 1.0220123031814914)
(150, 0.9783359728089053)
(160, 1.0108277345016037)
(170, 1.1087646700114135)
(180, 1.0097750809231012)
(190, 1.017119983666107)
(200, 1.0468622987085805)
    };
    \addlegendentry{Random span masking}
    \addplot[mark=o,g-orange] plot coordinates {
(0, 0.24583594144307278)
(10, 0.3688633470909383)
(20, 0.36925738236742056)
(30, 0.416591803198644)
(40, 0.4619641520176722)
(50, 0.4627946554077122)
(60, 0.5223345536397)
(70, 0.4888092925400017)
(80, 0.5354503663593189)
(90, 0.5411899211868447)
(100, 0.5492647498842971)
(110, 0.5604785949541389)
(120, 0.556891506009893)
(130, 0.5717637951899287)
(140, 0.5974641177394396)
(150, 0.6012307466956258)
(160, 0.6146574663791348)
(170, 0.6005905914097749)
(180, 0.5742741544145618)
(190, 0.6144357218130096)
(200, 0.6187538774662582)
};
\addlegendentry{Random uniform masking}
\end{axis}
\end{tikzpicture}
\caption{The Retrieval Utility (RU, described in Eq.~\ref{eq:retrieval_utility}) vs the number of pre-training steps. RU roughly estimates the ``usefulness''
of retrieval. RU is impacted by the choice of masking and the number of pre-training steps. }
\label{fig:retrieval_utility}
\end{figure*}

The null document $\znull$ described in Section~\ref{sec:inductive-bias}
provides a way to measure the importance of a retrieved document $z$: we define the \emph{retrieval utility} (RU) of $z$ for the masked input $x$ as
the difference between the log-likelihood of the knowledge-augmented encoder when conditioning on $z$ versus on $\znull$:
\begin{equation}
\text{RU}(z \mid x) = \log p(y \mid z, x) - \log p(y \mid \znull, x). 
\label{eq:retrieval_utility}
\end{equation}
A negative RU shows that $z$ is less useful for predicting $y$ than the null document. This could mean that $z$ is irrelevant to $x$, but could also mean that the masked tokens in $x$ do not require world knowledge to predict, or that the world knowledge is sufficiently commonplace it has been baked into the model's parameters. In practice, we find that RU increases steadily over the course of pre-training, and is more predictive of good performance on the downstream task of \openqa than even the overall log-likelihood. An example of how RU behaves over time and across different settings is in Figure~\ref{fig:retrieval_utility}.
   \bibliographystyle{icml2020}
\end{document}